# Towards Measuring Place Function Similarity at Fine Spatial Granularity with Trajectory Embedding

Cheng Fu, Robert Weibel

Department of Geography, University of Zurich

Contact information: Phone: 41-791213041, Email: cheng.fu@geo.uzh.ch

**ABSTRACT:** Modeling place functions from a computational perspective is a prevalent research topic. Trajectory embedding, as a neural-network-backed dimension reduction technology, allows the possibility to put places with similar social functions at close locations in the embedding space if the places share similar chronological context as part of a trajectory. The embedding similarity was previously proposed as a new metric for measuring the similarity of place functions. This study explores if this approach is meaningful for geographical units at a much smaller geographical granularity compared to previous studies. In addition, this study investigates if the geographical distance can influence the embedding similarity. The empirical evaluations based on a big vehicle trajectory data set confirm that the embedding similarity can be a metric proxy for place functions. However, the results also show that the embedding similarity is still bounded by the distance at the local scale.

## 1. INTRODUCTION

### 1.1. BACKGROUND

Knowing the social function of places over an urban area at different spatial granularities is essential for understanding the urban dynamics. Place functions also serve as fundamental inputs for domain applications such as modeling urban carbon emission [1], [2], hazard resilience [3], transportation modeling [4], or predicting the next place to visit [5]. The function of places, such as the amenity, socio-economic function or activities afforded by a place, or its land use, are commonly surveyed and labeled by predefined hierarchical taxonomies, for instance, in the form of commercial or user-contributed points-of-interest (POIs), or modeled by the spatio-temporal activity patterns *per se* [6]–[8]. 'Per se' in this case denotes that the place function is evaluated solely based on properties of the place itself, rather than by its spatial context, or functional relationships to other places. Recently, there has been a call for understanding places by relations [9]. Embedding models, and Word2Vec [10] in particular, have also been adapted as a useful prototype for modeling complex geographical or behavioral contextual information. Originally, the Word2Vec model aims to model a word's position in an embedding space by its preceding and succeeding words in sentences of a text corpus. In the embedding space, words with similar semantics are expected to have closer distances. One main application of the embedding modeling in GIScience is to model the similarity of POI categories, such as, whether cafes are more similar to restaurants than to hotels in general. The Place2Vec model [11] uses the neighboring POIs of the target POI as the spatial context for measuring the similarity of POI categories. A more recent perspective is to model the similarity of individual POIs. The Move2Vec model [12] takes the consecutive places in a movement as the context. Both [12] and [13] proposed that the function of places can be inferred by their preceding and succeeding neighbors in the trajectories but left it as an open question.

This study continues the conceptual model by [12] but focuses on investigating if the embedding space can measure place function similarity. Firstly, we apply the Word2Vec model to place-based trajectories at a much finer spatial granularity compared to previous studies. We then use POIs as a reference for validating the results. Particularly, we explore if the similarity in the embedding space indicates the similarity of place functions. We also explore if spatial proximity still has an impact on the output of this place modeling method, even though no explicit spatial information is involved in the processing.

## 1.2. RELATED WORK AND RESEARCH GAPS

One main approach of applying the embedding models to place modeling is to understand the similarity of POI categories, which we term *spatial context embedding*. Typically, POIs with the same category in an existing POI data set are treated as an identical word, and neighboring POIs within a buffering distance are used as the context to feed into the Word2Vec model or its variants. The embedding space then learns whether one POI category is more similar to another POI category ([11], [14]–[16]). This first approach requires places being already labeled, which may be limited by the potential spatial-temporal bias of the employed POI database [17]. In addition, this approach assumes that two POIs with the same category should have the same social function. However, even the same categories of POIs may have very different activity patterns. There has been a research agenda that also proposes to model each place as unique entity using spatial-context embedding [18].

A second, alternative approach advocated in this work is to treat each place as a unique entity, and trajectories formed by the places are fed into Word2Vec or its variants, which we term *trajectory embedding*. The preceding and succeeding places in the same trajectory thus are used as the context of a place for embedding. [19] uses the similarity of stay points learned by embedding from users' GPS trajectories as features for classifying demographic profiles of users. However, that particular study uses the embedding space of the places as a black box for feature engineering but does not explore the meanings of the embedding results in geography. [12] demonstrates on some examples, using call detail records (CDRs) as the input data, that the similarity of places in the embedding space, which we term *embedding similarity*, is associated with the similarity of social functions. However, as [12] uses the service area of cellular towers that have a 2-km minimal granularity as the geometry of places, the places correspond to very different geographical granularities, leading to a mixture of social functions represented in a particular cell, due to the nature of the spatial layout of the cellular towers.

## 1.3. RESEARCH QUESTIONS

Previous studies have shown initial insights that the trajectory embedding has the potential to model the social function of places by movements. The embedding similarity of individual places, then, has the potential to measure the similarity of social functions. In this study, we would like to address two additional questions based on extending previous studies:

**RQ1**: Can the similarity in the trajectory embedding space measure the similarity of social functions for individual places at a fine geographical granularity?

**RQ2**: Does the spatial proximity, particularly the distance decay, still influence the embedding similarity?

## 2. DATA

Two main data sets were used. The GPS trajectory data set is provided by a fleet management service company based in Greece. The data set has GPS-waypoint trajectories of 5389 vehicles over one year from June 1$^{st}$, 2017 to June 30$^{th}$, 2018. The majority of the vehicles are professional vehicles, such as trucks, vans, and buses. The trajectories are mainly located in Greece, but also cover continental Europe. The second data set is a POI data set consisting of OpenStreetMap point features with their amenity labels covered by the top-three categories in the Geofabrik taxonomy [20]: *places of worship*, *POI* (mainly referring to commercial POIs), and *transport and traffic*, containing 7,600,000 objects in total. This POI data set is used as a reference for validating the results from the embedding method.

## 3. METHODOLOGY

The methodology of this study consists of two main stages: the place modeling (Figure 1) and the analytics performed on the outputs of the place modeling (Figure 2). In the first stage, the waypoint records of a vehicle over a day are identified as an independent trajectory. Raw GPS trajectories are converted to stop-move semantic trajectories, as the stops are usually associated with activities interacting with places. Only the waypoints identified as stops are retained after applying a stop-detection algorithm that can deal with noisy waypoints [21]. The stop waypoints of a trajectory then are geocoded to a 30-m regular grid tessellation. The cells are then coded by Morton coding [22] to make sure that each cell has a unique ID. If consecutive stop waypoints are located in the same cell, only one cell is used for representing these waypoints. After the preprocessing, each GPS waypoint trajectory is then converted to a sequence of cell IDs. As an analog, each cell is understood as a unique word, and a trajectory as a sentence.

Essentially, Word2Vec is a two-layer neural network that aims to represent a word by a numeric vector after a process either using a target word to predict its neighboring words in a sentence, i.e., the *skip-gram* approach, or using the neighboring words to predict a target word, i.e., the *continues-bag-of-word* (CBOW) approach. The learned vectors then form an embedding space of the words. Therefore, the converted trajectories can also be input to the Word2Vec model. In this empirical evaluation, the skip-gram approach is used for prediction in the Word2Vec model. Cosine similarity is used for measuring the embedding similarity of the cells.

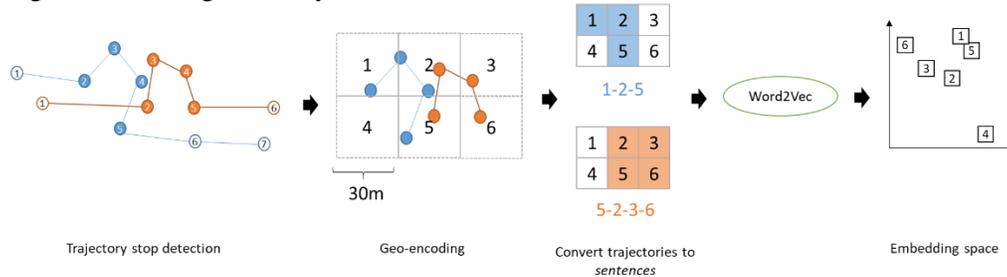

Figure 1 Overview of the place modeling workflow

In the second stage of analytics, the OSM POIs are associated to the same 30-m grid tessellation as the waypoints. If there is only one POI in a cell, the label of the cell is assigned as the category of the POI. If there is more than one POI contained in the same cell, the cell is labeled as *mixed*.

To answer RQ1 (Figure 2), we explore the embedding outputs by two tasks: Task 1 explores if an individual cell and its nearest neighbors share the same POI labels by geo-visual analytics. The selected cell and its most similar cells are visualized and investigated on the map. Task 2 explores if the cells with the same POI category are also close in the embedding space. The pairwise embedding similarities of cells with the same POI category are calculated. The embedding similarities of the selected cells and other types of POIs are also calculated. A two-sided t-test is applied to assess if the two embedding similarity sets are significantly different. In general, we shall expect that the intra-category similarity should be higher than the inter-category similarity if the embedding similarity is indeed able to measure the similarity of social functions.

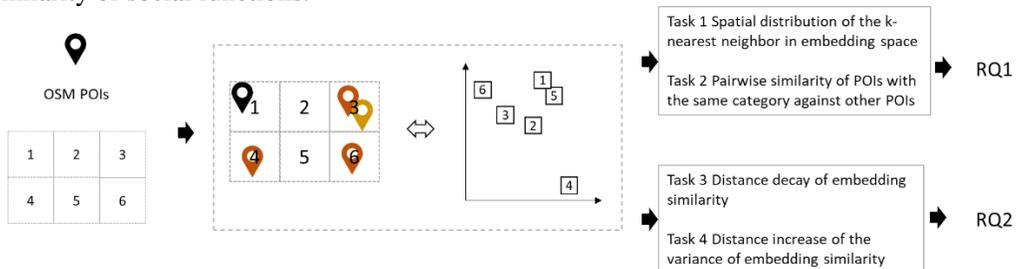

Figure 2 The framework of analytics

To answer RQ2 (Figure 2), we also perform two tasks. Task 3 is a quantitative approach to explore how the pairwise embedding similarity of the cells, both overall and by POI categories, may decay by geographical distance using linear regression fitting. Task 4 explores how the pairwise embedding similarity of the cells may vary by distance using an empirical semi-variogram. In the empirical evaluation, Euclidean distance is used to represent the geographical distance.

## 4. EVALUATION AND RESULTS

For the stop-detection setting, a stop is defined to last more than 5 minutes. 960,322 (14.7% out of all) 30-m cells were then detected containing at least one stop. Trajectories were composed of 6.6 cells containing stops on average (stddev = 5.2). For the Word2Vec model, the moving window size for defining a neighbor was set to 5. Cells with less than 5 overall visitations were excluded. The embedding vector size was set to 20. Eventually, 237,822 cells were modeled in the final embedding space, of which 14,745 were associated with at least one OSM POI.

As an example of visual analytics carried out in Task 1, we randomly selected a particular cell, i.e., a cell with a kiosk associated in Figure 3, which shows that its 10-nearest neighbors do share similar social functions with many target cells. For instance, *convenience* and *pharmacy* are similar to *kiosk*, and all belong to the retailing sector. Even the *park* cell in Figure 3.B actually has a convenience store and a pharmacy right next to it, as found by additional investigation in Google Maps. This suggests that trajectory embedding has the potential to identify POI location and categories that are not recorded in an existing POI database. It can also be observed that distance is not a dominant bound for this case as the cells sharing high cosine similarity are not nearby. However, we also observe some degree of locality because all the neighbors are located within the same city.

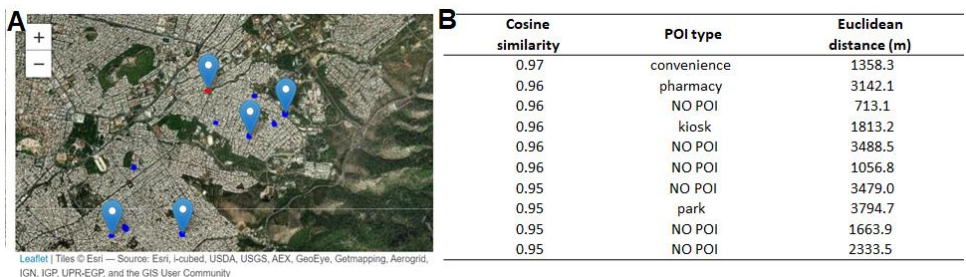

Figure 3 A randomly selected sample cell located near Athens, Greece with POI category *kiosk* and its ten most similar cells in the embedding space. A: The locations of the cells. The target *kiosk* cell is colored in red. Cells with a marker are those with a POI associated. B: The similarities of the cells, their associated POI categories, and their distances to the target cell. NO POI indicates the cell has no associated POI in the used OSM POI database.

For Task 2, it can be observed that cells with the same POI category are significantly more similar within the category than to other types of POIs, except for the fuel stations (
Table 1).

Table 1 The mean embedding similarity of POIs with the selected categories and the mean embedding similarity to other POIs. POI code and category is based on the Geofabrik taxonomy [20]. The default sample size for randomly selected POIs with the same category was 300, unless the total number of POIs was less than 300. The number of other POIs always matches the sample size of the selected POIs. (* with $p < 0.05$, *** with $p < 0.001$)

| POI code & category | Sample size | Intra-category mean similarity | Mean similarity to other POIs | T-test |
|---|---|---|---|---|
| 2101 pharmacy | 300 | 0.39 | 0.37 | 16.53*** |
| 2501 supermarket | 300 | 0.39 | 0.37 | 24.88*** |
| 2301 restaurant | 300 | 0.37 | 0.36 | 5.54*** |
| 2511 convenience | 300 | 0.39 | 0.37 | 12.19*** |
| 2562 car repair | 100 | 0.37 | 0.36 | 2.39* |
| 5250 fuel | 300 | 0.33 | 0.34 | 15.54*** |
| 2305 bar | 100 | 0.42 | 0.36 | 14.67*** |

|      |         |     |      |      |          |
|------|---------|-----|------|------|----------|
| 5260 | parking | 300 | 0.36 | 0.33 | 34.07*** |

For Task 3, 3,000 cells in the embedding space were randomly selected. The sampled cells are scattered over Europe, mainly in Greece, Italy, and France (Figure 4.A). Pairwise cosine similarities in the embedding space and Euclidean distances in geography were calculated, respectively. Overall, the cosine similarity decreases with increasing geographical distance (Figure 4.B), which fits the linear regression model $CS = -1.03 * 10^{-7} * D + 0.40$, where CS stands for the cosine similarity and D stands for the Euclidean distance. However, it can be observed that the trends are quite different at short range and at long range, respectively. For the cosine similarity within the local range (defined as D < 50,000 m), the distance decay is much stronger (Figure 4.C), which fits the model $CS = -5.48 * 10^{-6} * D + 0.64$. However, for the very long-range distance relationships, the cosine similarity actually increases with increasing distance (Figure 4.D), which roughly fits the model $CS = 3.02 * 10^{-7} * D - 0.62$. Therefore, it can be inferred that the similarity of the cells after the trajectory embedding is influenced by the distance and the amenity of the place at different scales. At the local level, the distance has a strong influence, as its slope is about 50 times larger than the slope of the overall regression model, although the absolute value is still small. The positive value of the model fitted for the long distances indicates that the embedding similarity does reflect the similarity of the social function of the places. This suggests evidence that two places far apart may still provide similar amenities, e.g., a restaurant in Paris, France and a restaurant in Athens, Greece both serve food, regardless of the geographical distance between them.

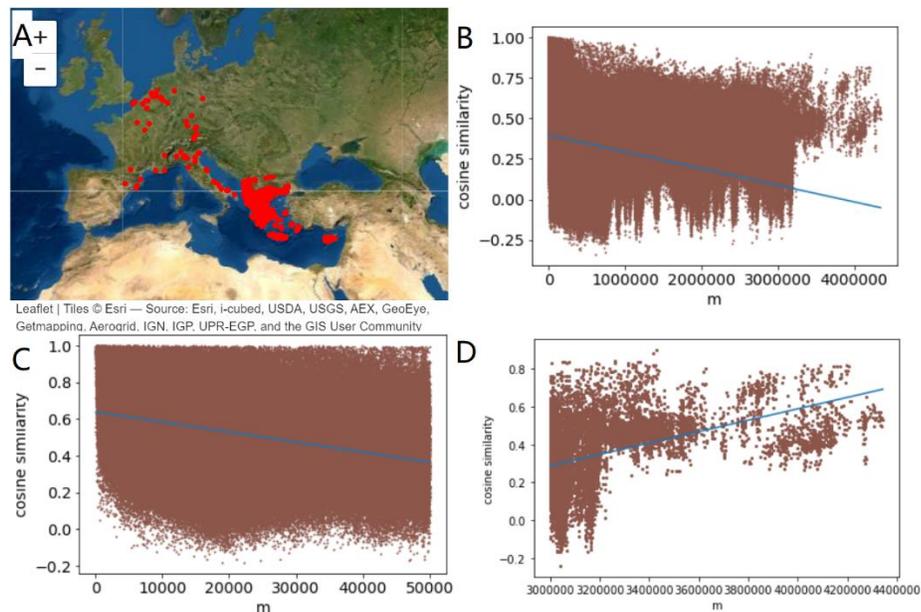

Figure 4 A: The pairwise cosine similarities against the corresponding pairwise Euclidean distances of the 3,000 sample cells. B: The locations of the 3,000 sample cells. C: The linear regression model fitted to the cosine similarity values with distances within 50,000 m. D: The model fitted to the cosine similarity values with distances larger than 3000 km.

For exploring the experimental semi-variograms, the same sample of 3,000 cells as used for Task 3 was employed (Figure 5.A). The bin bandwidth was set to 1000 m and the maximum distance as set as 100 km. In addition, the semi-variograms of a sample of 200 random cells with the desired POI categories (Figure 5.B-F) was taken. Unlike common physical geographical phenomena, whose semi-variance increases along with distance and reaches at a sill, it can be observed that the variances of cosine similarity are complex but stable in general, with very gently overall slopes ($\sim 10^{-7}$). The variances reach a local peak at short distances (< 50,000 m) for several POI categories, suggesting some local clusters. At long distances, the similar of social functions may contribute to the decrease or sill of the semi-variances. The local cluster of all POI categories (Figure 5.A) shows evidence of the influence of distance because vehicles usually have a limited activity space. The local clusters of the individual POI categories, however, indicate that place functions also contribute to the clustering effect.

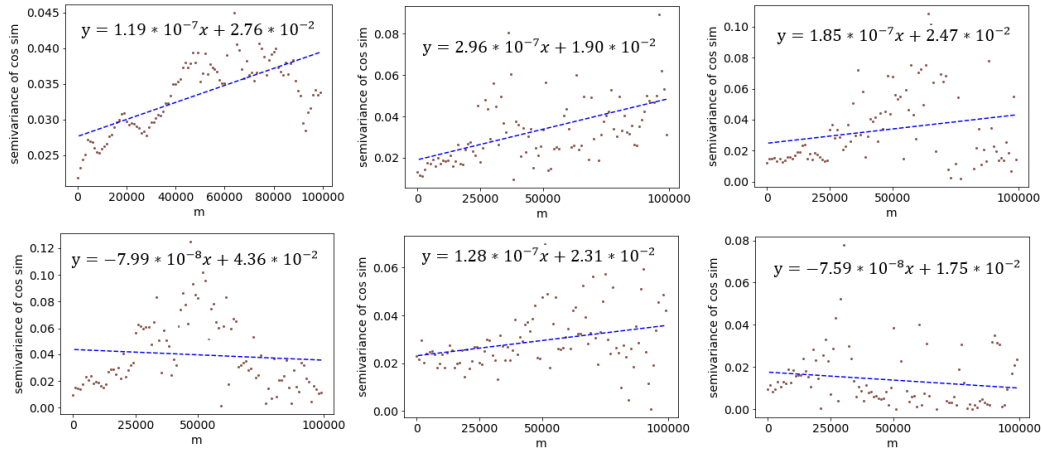

Figure 5 Variograms and corresponding linear fitted models of cosine similarity against pairwise distances for 3000 cells sampled randomly from all cells, and cells with selected POI categories (200 cells sampled for each POI category). A: all POI categories; B: *supermarket*; C: *restaurant*; D: *convenience*; E: *fuel*; F: *bar*.

## 5. CONCLUSIONS AND FUTURE WORK

Given the results of the preliminary evaluation by Task 1 and Task 2, RQ1 can be answered that the trajectory-based embedding result can reflect the similarity of social function in real life. It is observed that if cells have the same POI associated, they most likely have a higher embedding similarity as well. The First Law of Geography still matters to the embedding similarity based on the observation of outputs from Task 3 and Task 4, even though the process of trajectory embedding does not model the distance explicitly. A possible reason might be that most vehicles still have a relatively limited area to move that the embedding model still learned some local knowledge. However, a relative stability of variance against distance shows that social functions of the places do contribute to the cells' similarity in the embedding space.

In summary, the embedding similarity is both a result of locality and the similarity of social function. The advantage of embedding similarity as a metric is that it does not require prior hard-coded categories for POIs, yet may still provide semantically meaningful place recommendations given a source place as the preliminary workflow of Figure 1. This sheds light on recommending nearby and similar alternatives for a place required by a customer of a location-based service. We would like to explore this possibility further in future work.

## REFERENCES

[1] E. L. Glaeser and M. E. Kahn, "The greenness of cities: Carbon dioxide emissions and urban development," *J. Urban Econ.*, vol. 67, no. 3, pp. 404–418, May 2010.
[2] G. Wang, Q. Han, and B. de Vries, "A geographic carbon emission estimating framework on the city scale," *J. Clean. Prod.*, vol. 244, p. 118793, Jan. 2020.
[3] R. J. Burby, R. E. Deyle, D. R. Godschalk, and R. B. Olshansky, "Creating Hazard Resilient Communities through Land-Use Planning," *Nat. Hazards Rev.*, vol. 1, no. 2, pp. 99–106, May 2000.
[4] M. Iacono, D. Levinson, and A. El-Geneidy, "Models of Transportation and Land Use Change: A Guide to the Territory," *J. Plan. Lit.*, vol. 22, no. 4, pp. 323–340, May 2008.
[5] C. Schreckenberger, S. Beckmann, and C. Bartelt, "Next Place Prediction," in *Proceedings of the 2nd ACM SIGSPATIAL Workshop on Prediction of Human Mobility - PredictGIS 2018*, 2019, pp. 37–45.
[6] B. Huang, B. Zhao, and Y. Song, "Urban land-use mapping using a deep convolutional neural network with high spatial resolution multispectral remote sensing imagery," *Remote Sens. Environ.*, vol. 214, no. April, pp. 73–86, Sep. 2018.

[7] H. Xing, Y. Meng, and Y. Shi, "A dynamic human activity-driven model for mixed land use evaluation using social media data," *Trans. GIS*, vol. 22, no. 5, pp. 1130–1151, 2018.

[8] A. R. Bahrehdar, B. Adams, and R. S. Purves, "Streets of London: Using Flickr and OpenStreetMap to build an interactive image of the city," *Comput. Environ. Urban Syst.*, vol. 84, no. December 2019, p. 101524, 2020.

[9] R. S. Purves, S. Winter, and W. Kuhn, "Places in Information Science," *J. Assoc. Inf. Sci. Technol.*, vol. 70, no. 11, pp. 1173–1182, 2019.

[10] T. Mikolov, K. Chen, G. Corrado, and J. Dean, "Distributed Representations of Words and Phrases and their Compositionality," *Nips*, pp. 1–9, 2013.

[11] B. Yan, K. Janowicz, G. Mai, and S. Gao, "From ITDL to Place2Vec," in *Proceedings of the 25th ACM SIGSPATIAL International Conference on Advances in Geographic Information Systems - SIGSPATIAL'17*, 2017, pp. 1–10.

[12] A. Crivellari and E. Beinat, "From Motion Activity to Geo-Embeddings: Generating and Exploring Vector Representations of Locations, Traces and Visitors through Large-Scale Mobility Data," *ISPRS Int. J. Geo-Information*, vol. 8, no. 3, p. 134, 2019.

[13] S. Mehmood and M. Papagelis, "Learning Semantic Relationships of Geographical Areas based on Trajectories," in *2020 21st IEEE International Conference on Mobile Data Management (MDM)*, 2020, pp. 109–118.

[14] Y. Yao, X. Li, X. Liu, P. Liu, Z. Liang, and J. Zhang, "Sensing spatial distribution of urban land use by integrating points-of-interest and Google Word2Vec model," *Int. J. Geogr. Inf. Sci.*, vol. 31, no. 4, pp. 825–848, 2017.

[15] W. Zhai, X. Bai, Y. Shi, Y. Han, Z.-R. Peng, and C. Gu, "Beyond Word2vec: An approach for urban functional region extraction and identification by combining Place2vec and POIs," *Comput. Environ. Urban Syst.*, vol. 74, no. December 2018, pp. 1–12, Mar. 2019.

[16] K. Liu, L. Yin, F. Lu, and N. Mou, "Visualizing and exploring POI configurations of urban regions on POI-type semantic space," *Cities*, vol. 99, no. January, p. 102610, Apr. 2020.

[17] G. Mckenzie, K. Janowicz, C. Keßler, and G. Mckenzie, "Uncovering spatiotemporal biases in place-based social sensing," pp. 1–18, 2020.

[18] G. Giannopoulos and M. Meimaris, "Learning Domain Driven and Semantically Enriched Embeddings for POI Classification," in *Proceedings of the 16th International Symposium on Spatial and Temporal Databases*, 2019, pp. 214–217.

[19] A. Solomon, A. Bar, C. Yanai, B. Shapira, and L. Rokach, "Predict Demographic Information Using Word2vec on Spatial Trajectories," in *Proceedings of the 26th Conference on User Modeling, Adaptation and Personalization - UMAP '18*, 2018, pp. 331–339.

[20] F. Ramm, "OpenStreetMap Data in Layered GIS Format," 2017.

[21] L. Xiang, M. Gao, and T. Wu, "Extracting stops from noisy trajectories: A sequence oriented clustering approach," *ISPRS Int. J. Geo-Information*, vol. 5, no. 3, 2016.

[22] G. M. Morton, "A computer oriented geodetic data base and a new technique in file sequencing," Ottawa, Canada, 1966.